\documentclass[letterpaper]{article} 
\usepackage{aaai2026}  
\usepackage{times}  
\usepackage{helvet}  
\usepackage{courier}  
\usepackage[hyphens]{url}  
\usepackage{graphicx} 
\usepackage{natbib}  
\usepackage{caption} 
\usepackage{epstopdf}
\usepackage{multirow}
\usepackage{color}
\usepackage[table]{xcolor}
\usepackage{colortbl}
\usepackage{array}
\usepackage{arydshln}
\usepackage{bm}
\usepackage{tabularx}
\usepackage{amssymb}
\usepackage{float}
\usepackage[linesnumbered,ruled,vlined]{algorithm2e}
\SetKwInput{KwIn}{Input}
\SetKwInput{KwOut}{Output}
\usepackage{float}

\usepackage{soul}
\usepackage{url}
\usepackage[utf8]{inputenc}
\usepackage{amsmath, amssymb}

\usepackage{booktabs}
\usepackage[switch]{lineno}

\usepackage{colortbl} 
\definecolor{tabletitle}{gray}{.8}
\definecolor{ours}{gray}{.95}
\definecolor{ggray}{RGB}{127,127,127}
\definecolor{reda}{RGB}{202,0,0}
\definecolor{redb}{RGB}{217,148,143}
\definecolor{myyellow}{RGB}{190,144,0}
\definecolor{mygreen}{RGB}{0,136,51}
\definecolor{myblue}{RGB}{0,102,204}
\definecolor{myPurple}{RGB}{117,107,177}
\definecolor{myOrange}{RGB}{230,85,13}
\definecolor{myTeal}{RGB}{27,158,119}
\newcolumntype{B}{!{\vrule width 1pt}}
\usepackage{pifont}

\usepackage{newfloat}
\usepackage{listings}
\DeclareCaptionStyle{ruled}{labelfont=normalfont,labelsep=colon,strut=off} 
\floatstyle{ruled}
\newfloat{listing}{tb}{lst}{}
\floatname{listing}{Listing}
\title{Small but Mighty: Dynamic Wavelet Expert-Guided Fine-Tuning of Large-Scale Models for Optical Remote Sensing Object Segmentation}

\author {
    Yanguang Sun\textsuperscript{\rm 1},
    Chao Wang\textsuperscript{\rm 1},
    Jian Yang\textsuperscript{\rm 2},
    Lei Luo\textsuperscript{\rm 1} \thanks{Corresponding author.}
}
\affiliations {
    \textsuperscript{\rm 1}PCA Lab, Nanjing University of Science and Technology, Nanjing, China\\
    \textsuperscript{\rm 2}PCA Lab, VCIP, College of Computer Science, Nankai University, Tianjin, China\\
    {Sunyg@njust.edu.cn, wchao0601@163.com, csjyang@nankai.edu.cn,  luoleipitt@gmail.com}
}

\usepackage{bibentry}

\begin{document}

\maketitle

\begin{abstract}
Accurately localizing and segmenting relevant objects from optical remote sensing images (ORSIs) is critical for advancing remote sensing applications. Existing methods are typically built upon moderate-scale pre-trained models and employ diverse optimization strategies to achieve promising performance under full-parameter fine-tuning. In fact, deeper and larger-scale foundation models can provide stronger support for performance improvement. However, due to their massive number of parameters, directly adopting full-parameter fine-tuning leads to pronounced training difficulties, such as excessive GPU memory consumption and high computational costs, which result in extremely limited exploration of large-scale models in existing works. In this paper, we propose a novel dynamic wavelet expert-guided fine-tuning paradigm with fewer trainable parameters, dubbed WEFT, which efficiently adapts large-scale foundation models to ORSIs segmentation tasks by leveraging the guidance of wavelet experts. Specifically, we introduce a task-specific wavelet expert extractor to model wavelet experts from different perspectives and dynamically regulate their outputs, thereby generating trainable features enriched with task-specific information for subsequent fine-tuning. Furthermore, we construct an expert-guided conditional adapter that first enhances the fine-grained perception of frozen features for specific tasks by injecting trainable features, and then iteratively updates the information of both types of feature, allowing for efficient fine-tuning. Extensive experiments show that our WEFT not only outperforms 21 state-of-the-art (SOTA) methods on three ORSIs datasets, but also achieves optimal results in camouflage, natural, and medical scenarios. The source code is available at: https://github.com/CSYSI/WEFT
\end{abstract}

\section{Introduction}
In recent years, object segmentation in optical remote sensing images (ORSIs) has become a research focus \cite{GeleNet,SFANet,DPU-Former}, owing to its broad applications in areas such as urban planning, agricultural monitoring, disaster assessment, and military reconnaissance. Unlike common natural images \nocite{RCNet,DSP,LPMoE,GLCONet,EMCENet,DMINet}, ORSIs are typically captured by sensors mounted on aircraft or satellites and are presented as bird's-eye views. As a result, targets in ORSIs \nocite{Wang1} often exhibit arbitrary orientations, drastic scale variations, and dense distributions across complex backgrounds, making their segmentation particularly challenging. 

\begin{figure}[t]
      \centering\includegraphics[width=0.48\textwidth,height=7cm]{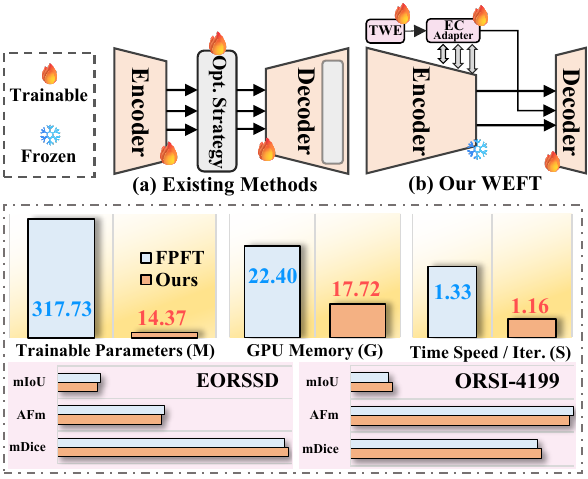}
      \captionsetup{font={small}}
      \caption{The top part shows architectural differences between existing methods in ORSIs segmentation tasks and our WEFT framework. The bottom part presents a comparison of efficiency and performance between full-parameter fine-tuning (FPFT) and the proposed WEFT method, conducted under the same architecture.}
	\label{FIG1}
\end{figure}

To achieve promising performance, a large number of deep learning–based ORSIs object segmentation methods \cite{ORSSD,MCCNet,ESGNet,SFANet} have been proposed one after another. At the beginning, the architectures (as depicted in Fig. \ref{FIG1}(a)) of these methods are primarily based on pre-trained convolutional encoders ($e.g.$, VGG16 \cite{VGG16}, or ResNet50 \cite{ResNet}), which extract initial features that are then enhanced through sophisticated optimization strategies \cite{MCCNet,ACCorNet,SRAL,SFANet,LIAN1}. Although these methods achieve commendable performance by updating all parameters, the inherent limitation of convolutional networks ($i.e.$, local receptive fields), makes it difficult to further improve performance. Soon after that, Transformer architectures \cite{ViT, Pvt2} rose to prominence due to their superior modeling capacity and effectively mitigated this limitation.

Building on this, numerous Transformer-based methods \cite{GeleNet,TLKCDNet,UDCNet,ADSTNet,DPU-Former} are introduced and achieve better performance in ORSIs object segmentation tasks. 
This performance advantage stems in part from global modeling, with another important factor being the model scale. For example, Transformer-based architectures ($e.g.$, Swin-B \cite{Swin}, 87.77M parameters; PVTv2-B4 \cite{Pvt2}, 62.60M; and DPU-Former \cite{DPU-Former}, 38.89M) used in existing methods \cite{ICON-p, UDCNet, DPU-Former} typically have two to three times more trainable parameters than their convolutional counterparts, or even more. Indeed, the scale of the above architectures is not particularly large, but rather moderate. However, larger-scale and deeper foundation models \cite{uniper, dinov2} have been explored in current architecture research and have demonstrated outstanding performance across visual tasks \nocite{Lian2,YU1,YU2}. 

Inspired by this, we believe that rather than relying on sophisticated strategies to improve performance, it is more effective to embed a deeper and larger-scale foundation model to encode strong discriminative features from input images. With this goal in mind, we abandon small- and moderate-scale foundation models in terms of framework design and instead introduce a deeper, larger-scale, and more parameter-rich foundation model ($e.g.$, UniPerceiver-L \cite{uniper} with 303.36M parameters). At first, we adopt the same training strategy ($i.e.$, full-parameter fine-tuning (FPFT)) as existing works \cite{GeleNet,SFANet,DPU-Former} to optimize the model. However, in Fig. \ref{FIG1}, we find that this strategy struggles to cope with large-scale foundation models with massive parameters, as all parameters must be updated simultaneously during training. This significantly increases the computational burden during both the forward and backward passes and leads to a sharp increase in GPU memory consumption, making the optimization process especially difficult. In particular, when dealing with high-resolution inputs or large batch sizes, memory usage often approaches hardware limits, easily resulting in resource bottlenecks or even training interruption. This constitutes an important reason for the relatively limited exploration of large-scale models in ORSIs segmentation tasks. 

Considering this key factor, we no longer use the typical FPFT strategy to train the network, and instead propose a novel dynamic wavelet expert-guided fine-tuning paradigm, termed WEFT. As shown in Fig. \ref{FIG1}(b), we keep the large-scale model frozen with its parameters unchanged, and introduce a lightweight, trainable branch in parallel to supplement task-specific information that the frozen branch lacks, thereby enabling more effective adaptation to ORSIs tasks. From Fig. \ref{FIG1}, the performance of our WEFT is almost the same as that under full-parameter fine-tuning, but our WEFT only requires 14.37M trainable parameters, which only accounts for 4.52\% of the entire framework parameters. Moreover, it reduces training GPU memory consumption by approximately 26.41\% and improves training speed (1 iter.) by 14.66\%. Specifically, our WEFT framework consists of two lightweight components: the task-specific wavelet expert (TWE) extractor and the expert-guided conditional (EC) adapter. Technically, the TWE extractor adaptively integrates wavelet experts from different perspectives to obtain trainable features rich in task-oriented knowledge from input images. Subsequently, the EC adapter is used to supplement task-specific details in frozen features by merging trainable features, and further strengthens both types of features, enabling efficient fine-tuning. Extensive experiments on three ORSIs benchmark datasets demonstrate that our WEFT not only clearly surpasses 21 state-of-the-art (SOTA) approaches but also requires far few trainable parameters. Additionally, it exhibits strong generalization capabilities in camouflage, natural, and medical scenarios. 

The main contributions can be summarized as follows:

$\bullet$ A novel dynamic wavelet expert-guided fine-tuning (WEFT) is proposed to efficiently adapt large-scale foundation models to object segmentation tasks in ORSIs.

$\bullet$ A lightweight task-specific wavelet expert (TWE) extractor is introduced to obtain discriminative trainable features by adaptively combining various wavelet experts.

$\bullet$ An efficient expert-guided conditional (EC) adapter is designed to reconstruct the internal information within trainable and frozen features through conditional optimization.

\section{Related work}
\textbf{Optical remote sensing object segmentation.} The primary objective of ORSIs object segmentation tasks is to segment meaningful remote sensing targets. Initially, convolution-based models \cite{ORSSD, MCCNet} are widely proposed, with various optimization strategies ($e.g.$, attention \cite{EORSSD}, boundary auxiliary \cite{ERPNet, ESGNet}, and multi-scale enhancement \cite{SFANet, LGIPNet}) designed to optimize initial features, yielding promising results. However, due to the limited local perception of convolutional encoders \cite{ResNet}, these methods face performance bottlenecks. Considering this factor, GeleNet \cite{GeleNet}, TLCKDNet \cite{TLKCDNet}, and DPU-Former \cite{DPU-Former} introduced moderate-scale Transformer-based encoders (such as PVTv2 \cite{PVT}) in their design and achieve excellent accuracy by updating all parameters. We believe that architectures based on large-scale models ($e.g.$, UniPerceiver \cite{uniper}) are more advantageous for tackling challenging ORSIs tasks. However, the FPFT strategy adopted by existing works struggles to deploy these models effectively, primarily due to their massive parameters.
\begin{figure*}[t]
      \centering\includegraphics[width=0.99\textwidth,height=3.35cm]{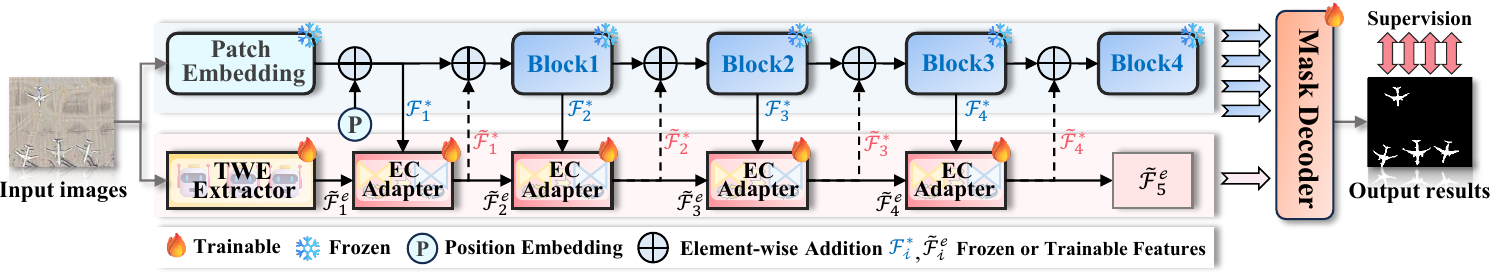}
      \captionsetup{font={small}}
      \caption{Overview of our WEFT framework. The overall architecture comprises a frozen foundation model \cite{uniper}, a Task-specific Wavelet Expert (TWE) extractor, four Expert-guided Conditional (EC) adapters, and a mask decoder \cite{mask2}. In the training process, the proposed WEFT method requires only 14.37M trainable parameters while achieving excellent performance.}
	\label{liuchengtu}
\end{figure*}

\noindent \textbf{Fine-tuning of large-scale foundation models.} Large-scale models \cite{uniper, dinov2} refer to foundation architectures with deep structures and large number of parameters, pre-trained on massive datasets. Their powerful modeling capabilities provide strong support for achieving excellent performance in downstream tasks. However, the application of these models is often challenging due to their enormous parameters, which leads to significant computational overhead and makes training difficult when using the FPFT strategy. Recently, several fine-tuning methods \cite{LORA,VPT} have been proposed to adapt large-scale models to visual tasks. Specifically, LoRA \cite{LORA} achieved parameter fine-tuning by inserting low-rank trainable matrices into pre-trained models. VPT \cite{VPT} embedded some small learnable prompts in a frozen backbone to allow adaptation to recognition tasks.

Unlike the above methods, we propose an innovative fine-tuning paradigm based on dynamic wavelet expert guidance to alleviate the training difficulties encountered when applying large-scale models to ORSIs object segmentation tasks. 

\section{Methodology}

\subsection{Overall Framework}
The overall framework of our WEFT is illustrated in Fig. \ref{liuchengtu}, which consists of four key components: \textbf{(a)} UniPerceiver-L \cite{uniper} foundation model with frozen parameters, \textbf{(b)} Task-specific wavelet expert (TWE) extractor, \textbf{(c)} Expert-guided conditional (EC) adapter, and \textbf{(d)} mask decoder \cite{mask2}. Specifically, for an input image $\mathcal{I}_m\in \mathbb{R}^{3\times H \times W}$, we adopt a dual-branch architecture ($i.e.$, the frozen UniPerceiver-L network and the trainable TWE extractor) to simultaneously extract frozen and trainable features. Subsequently, the trainable features $\{{\mathcal{F}}_i^{\diamond}\}_{i=1}^{4}$ and the frozen features $\{{\mathcal{F}}_i^{\ast}\}_{i=1}^{5}$ are fed into the EC adapter, where information guidance and reinforcement are achieved through collaborative operations of deformable attention \cite{Deform}, edge-aware subspace token optimizer (ESTO), and spatial-aware expert enhancer (SEE). Ultimately, the optimized features are input into a lightweight mask decoder to generate the final segmentation results.

\subsection{Task-specific Wavelet Expert Extractor}
The task‑specific wavelet expert (TWE) extractor is designed to dynamically acquire wavelet experts that are rich in task‑related knowledge through well-designed wavelet convolutions \cite{WC}, thereby providing strong informational support for subsequent fine‑tuning. Not only does it provide task‑specific information to the frozen foundation model, but it also supplements local details within the Transformer structure, refining object boundaries.

\begin{figure}[t]
      \centering\includegraphics[width=0.47\textwidth,height=4.8cm]{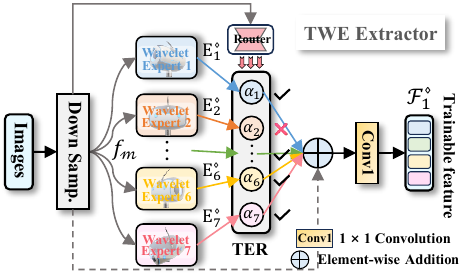}
      \captionsetup{font={small}}
      \caption{Detailed diagram of the first stage in TWE extractor.}
	\label{module1}
\end{figure}

Specifically, as shown in Fig. \ref{module1}, we first perform a down-sampling ($\mathsf{DS}(\cdot)$) on the image $\mathcal{I}_m$, $i.e.$, $f_m = \mathsf{DS}(\mathcal{I}_m)$, and then model various wavelet experts $\{\mathrm{E}_n^{\diamond}\}_{n=1}^{7}$, which are obtained through wavelet convolutions \cite{WC} from different perspectives and enriched with task-specific knowledge. Unlike standard convolutions, wavelet convolution is more lightweight and increases diversity by modeling features from four directions ($i.e.$, HH, HL, LH, and LL). This entire process can be formulated as follows:
\begin{equation}
\begin{split}
        &\{\mathrm{E}_n^{\diamond}\}_{n=1}^{7}=\mathcal{C}_1(\mathsf{Cat}[\tilde{f}_{n}^{1},\tilde{f}_{n}^{2},\tilde{f}_{n}^{3},\tilde{f}_{n}^{4}])+f_m, \\
        &\tilde{f}_{n}^{k}=\mathcal{WC}_{2n-1}(f_{n}^{k}+\tilde{f}_{n}^{k-1}), \{f_{n}^{k}\}_{k=1}^{4}=\mathsf{Split}(f_m),\\
        & \mathcal{WC}_{2n-1}(x)=\mathsf{IWT}(\mathcal{DC}_{2n-1}(w_{c},x_{hh},x_{hl},x_{lh},x_{ll}) ),
\end{split}
\end{equation}
where $\mathcal{C}_1(\cdot)$ is a 1$\times$1 convolution, $\mathsf{Cat}[\cdot]$ and ``$+$'' present concatenation and element-wise addition, $\mathcal{WC}_{2n-1}(\cdot)$ and $\mathcal{DC}_{2n-1}(\cdot)$ represent a wavelet convolution and a depthwise convolution with kernel size 2$n$-1$\times$2$n$-1, $\mathsf{Split}(\cdot)$ is a separation operation that divides the feature into four sub-features along the channels. $\mathsf{IWT}(\cdot)$ is the inverse wavelet transform. ``$(x_{hh}, x_{hl}, x_{lh}, x_{ll})$'' are obtained from the wavelet transform of $x$, denoted as ``$\mathsf{WT}(x)$'', ``$w_c$'' is a weight tensor. Based on this, we obtain multiple wavelet experts $\{\mathrm{E}_n^{\diamond}\}_{n=1}^{7}$ with varying receptive fields. Due to differences in their receptive fields, these experts carry distinct task-relevant knowledge, each specializing in object information at a specific scale. In ORSIs, the scale variation of targets is often significant. We argue that not all wavelet experts $\{\mathrm{E}_n^{\diamond}\}_{n=1}^{7}$ are equally beneficial. For small objects, an excessively large receptive field may introduce ambiguity, whereas for large objects, an overly small receptive field may lead to incomplete comprehension. Only those with appropriate receptive fields are effective in contributing to model fine-tuning. 

Therefore, we propose a \textbf{Top-k Expert Router (TER)}, which aims to dynamically select appropriate wavelet experts $\{\mathrm{E}_n^{\diamond}\}_{n=1}^{7}$ to achieve optimal matching of remote sensing targets with varying types and scales. Technically, we first obtain a set of learnable weights $\alpha$=\{$\alpha_1$, $\alpha_2$, ... , $\alpha_7$\} based on the input feature $f_m$, as shown in: 
\begin{equation}
\begin{split}
        &\alpha = \mathbf{Sof}(w_g(\frac{1}{HW}\sum_{i=1}^{H}\sum_{j=1}^{W}f_m[:,:,i,j])+b_g), \\
\end{split}
\end{equation}
where $\mathbf{Sof}(\cdot)$ presents the Softmax, ``$w_g$'' and ``$b_g$'' denote the learned weight tensor and bias vector from a linear layer, ``$(i,j)$'' are indices over the spatial dimension. Furthermore, we select the top-4 wavelet experts indices with the highest scores from the 7 weights generated by the gating network to form the set $\mathcal{T}$, $i.e.$, $\mathcal{T}=\mathbf{TopK}(\alpha, 4)\subset\{1,2,...,7\}, \left | \mathcal{T} \right|$ = 4, where $\mathbf{TopK}(\cdot)$ is an operator that selects the indices of the $k$ largest values from a given vector. The corresponding weights are then normalized to ensure a valid probability distribution, which is used as the final fusion coefficients $\tilde{\alpha}$=\{$\tilde{\alpha}_1$, $\tilde{\alpha}_2$, $\tilde{\alpha}_3$, $\tilde{\alpha}_4$\}. These coefficients are applied to the outputs of the selected wavelet experts to generate the final trainable feature $\tilde{\mathcal{F}}_1^{\diamond}$ via a weighted sum:
\begin{equation}
\begin{split}
        &{\mathcal{F}}_1^{\diamond}=\mathcal{C}_1(f_m + \sum_{u\in \mathcal{T}}\tilde{\alpha}_u \cdot \mathrm{E}_u^{\diamond}), \: \tilde{\alpha}_u = \frac{\mathsf{exp}({\alpha}_u)}{ {\textstyle \sum_{v\in \mathcal{T}}\mathsf{exp}({\alpha}_v)}},\\
\end{split}
\end{equation}
where ``$\cdot$'' is an element-wise multiplication, ``$u$'' denotes the index of a selected expert in the top-$k$ set $\mathcal{T}$, and ``$v$'' is used to compute the denominator over the selected experts.

In the subsequent process, the proposed TWE extractor adopts a classical hierarchical structure, which takes the generated trainable feature ${{\mathcal{F}}}_1^{\diamond}$ as input and progressively produces three multi-scale features \{${{\mathcal{F}}}_2^{\diamond}$, ${{\mathcal{F}}}_3^{\diamond}$, ${{\mathcal{F}}}_4^{\diamond}$\} enriched with task-related information through identical operations.

\subsection{Expert-guided Conditional Adapter}
The expert-guided conditional (EC) adapter is constructed to complement the task-oriented details of frozen features $\{{{\mathcal{F}}}_i^{\ast}\}_{i=1}^{5}$ $({{\mathcal{F}}}_i^{\ast}$ $\in$ $\mathbb{R}^{\frac{HW}{16^{2}}\times C})$ through trainable features $\{{\mathcal{F}}_i^{\diamond}\}_{i=1}^{4}$ $({\mathcal{F}}_i^{\diamond}$ $\in$ $\mathbb{R}^{\frac{H}{2^{i+1}}\times\frac{W}{2^{i+1}}\times C})$ enriched with expert knowledge, and to enable iterative updates between the trainable and frozen features. Specifically, for trainable features $\{{\mathcal{F}}_i^{\diamond}\}_{i=1}^{4}$ based on wavelet experts, we select three scale features with the same scale as and adjacent to the frozen feature for fine-tuning progress, $i.e.$, $\tilde{{\mathcal{F}}_1^{e}}$$=$ $\mathsf{Cat}[\mathsf{RS}({\mathcal{F}}_2^{\diamond}), \mathsf{RS}({\mathcal{F}}_3^{\diamond}), \mathsf{RS}({\mathcal{F}}_4^{\diamond})]$, where $\mathsf{RS}(\cdot)$ denotes a reshaping of the tensor, $\tilde{{\mathcal{F}}_1^{e}}$ $\in$ $ \mathbb{R}^{(\frac{HW}{8^{2}}+\frac{HW}{16^{2}}+\frac{HW}{32^{2}})\times C}$. Compared to a single scale, it contains more task-oriented knowledge. Technically, during the first stage fine-tuning, with $\tilde{{\mathcal{F}}_1^{e}}$ and $\mathcal{F}_1^{\ast}$ from the patch embedding as initial inputs, we first employ deformable attention \cite{Deform} to inject task-specific information into the frozen feature $\mathcal{F}_1^{\ast}$, as follows:
\begin{equation}
\begin{split}
        &\hat{\mathcal{F}}_1^{\ast}=DeformAttn(\mathbf{LN}(\mathcal{F}_1^{\ast}),\mathbf{LN}(\tilde{{\mathcal{F}}_1^{e}})),\\
\end{split}
\end{equation}
where $\mathbf{LN}(\cdot)$ denotes the layer normalization to stabilize the input distribution. Although deformable attention \cite{Deform} effectively injects task-specific knowledge, it focuses mainly on semantic alignment between feature sources, without explicitly modeling structural details within the fused representation. As a result, these tokens may suffer from semantic ambiguity and insufficient structural awareness, which can limit the accuracy of segmentation. 

\begin{algorithm}[t]
\caption{\small Edge-aware Subspace Token Optimizer}
\label{alg:edge_refine}

\KwIn{\small $\hat{\mathcal{F}}_1^{\ast} \in \mathbb{R}^{N \times C}$, subspaces $H>1$, temp. $\varrho$, weight $\lambda$}

Normalize $\hat{\mathcal{F}}_1^{\ast}$ along channel dimension\;
Split into $H$ subspaces: $\mathrm{Norm}(\hat{\mathcal{F}}_1^{\ast})\rightarrow \mathbf{F}_h~(h=1,\dots,H)$\;
Compute similarity: $\mathbf{S}_h \gets \mathbf{F}_h\mathbf{F}_h^{\top}/\sqrt{d}\cdot \varrho$\;
Attention: $\mathcal{A}_h \gets \mathrm{Softmax}(\mathbf{S}_h)$\;
Aggregate \& merge heads: $\mathbf{T}_1^{\ast} \gets [\mathcal{A}_h \cdot \mathbf{X}_h]_{h=1}^{H}$\;

Compute channel-wise variance: $\mathbf{V} \gets \mathrm{Var}(\hat{\mathcal{F}}_1^{\ast})$\;
Edge mask: $\mathbf{M} \gets \sigma\!\left((\mathbf{V}-\overline{\mathbf{V}})/(\sigma_{\mathbf{v}}+\epsilon)\right)$\;
Apply mask: $\widetilde{\mathbf{T}}_{1}^{\ast} \gets (\mathbf{1}+\lambda \mathbf{M}) \cdot \mathbf{T}_1^{\ast}$\;
Gate \& Residual: $\widetilde{\mathcal{F}}_1^{\ast} \gets \delta \cdot \widetilde{\mathbf{T}}_{1}^{\ast} + \hat{\mathcal{F}}_1^{\ast}$\;

\KwOut{optimized frozen features $\widetilde{\mathcal{F}}_1^{\ast} \in \mathbb{R}^{N \times C}$}
\end{algorithm}

Considering these problems, we design an \textbf{Edge-aware Subspace Token Optimizer (ESTO)}, which uses subspace-based token-to-token attention to model intra-feature relationships and incorporates an edge-aware modulation mechanism to further enhance structurally significant regions in tokens, as in Algorithm \ref{alg:edge_refine}. Technically, given the input feature $\hat{\mathcal{F}}_1^{\ast} \in \mathbb{R}^{N \times C}$, we first apply $L_2$ normalization along the channel dimension to stabilize the similarity computation across tokens, and then we reshape the normalized feature into $H$ subspaces, each with a dimension of $d = C / H$. To be specific, we obtain the $\mathbf{F}_h \in \mathbb{R}^{H \times N \times d}$ by reshaping the normalized feature into shape $(N, H, d)$ and transposing the first two dimensions. This subspace formulation enables each head to independently capture token-to-token relationships, providing diverse contextual perspectives across different parts of the feature space. For each subspace, we compute the pairwise similarity between tokens using the scaled dot product, and the formula can be expressed as:
\begin{equation}
\begin{split}
        &\mathbf{S}_h = \frac{\mathbf{F}_h({\mathbf{F}_h})^{\top}}{ \sqrt{d} \cdot \varrho }, \: \: \mathbf{S}_h \in \mathbb{R}^{H \times N \times N}, \\
\end{split}
\end{equation}
where ``$\top$'' denotes a transpose operation, ``$\sqrt{d}$ '' presents the scaling factor, ``$\varrho$'' is a temperature hyperparameter that adjusts the sharpness of the similarity distribution. The attention weights $\mathcal{A}_h$ are then obtained by applying softmax to the similarity matrix along the last dimension, $i.e.$, $\mathcal{A}_h=\mathbf{Sof}(\mathbf{S} _h)$. The attention weights are applied to the original (unnormalized) input feature $\hat{\mathcal{F}}_1^{\ast}$, reshaped into subspace form $\mathbf{X}_h \in \mathbb{R}^{H \times N \times d}$, to obtain refined subspace representations. These outputs from all subspaces are then concatenated to form the optimized token $\mathbf{T}_{1}^{\ast}$, as follows:
\begin{equation}
\begin{split}
        &\mathbf{T}_{1}^{\ast} = \mathsf{Cat}[\mathcal{A}_1\mathbf{X}_1,\mathcal{A}_2\mathbf{X}_2, ... \ ,\mathcal{A}_H\mathbf{X}_H] \in \mathbb{R}^{N \times C}. \\
\end{split}
\end{equation}
Furthermore, we introduce an edge-aware modulation mechanism based on channel-wise variance to refine structural details, particularly around object boundaries. We observe that tokens with higher variance typically correspond to more informative or structurally salient regions ($e.g.$, edges, and contours). Based on this observation, we estimate a soft edge-aware mask $\mathbf{M}$ using the channel-wise variance $\mathbf{V}$ of the optimized token $\mathbf{T}_{1}^{\ast}$. We then update the token representation using the obtained edge mask $\mathbf{M}$, formulated as:
\begin{equation}
\begin{split}
        & \: \widetilde{\mathbf{T}}_{1}^{\ast} = \left( \mathbf{1} + \lambda \cdot \mathbf{M} \right) \cdot \mathbf{T}_{1}^{\ast}, \: \mathbf{M} = \mathbf{Sig}\left( \frac{\mathbf{V} - \overline{\mathbf{V}}}{\sigma_{\mathbf{v}} + \epsilon} \right), \\
        &\mathbf{V} = \mathsf{Var}(\hat{\mathcal{F}}_1^{\ast}[n, :]) = \frac{1}{C} \sum_{c=1}^{C} \left( \hat{\mathcal{F}}_1^{\ast}[n, c] - \bar{f}_n \right)^2, 
\end{split}
\end{equation}
where ``$\mathbf{1}$'' is a matrix of all ones with the same dimensions as ``$\mathbf{M}$'', $\mathbf{Sig}(\cdot)$ is the Sigmoid function, ``$-$'' is the element-wise subtraction, $\overline{\mathbf{V}}$ and $\sigma_{\mathbf{v}}$ represent the mean and standard values of variance, $\lambda$ denotes the hyperparameter of modulating edge mask intensity, $\bar{f}_n$($\frac{1}{C} \sum_{c=1}^{C}\hat{\mathcal{F}}_1^{\ast}[n, c]$) present the mean value of the $n$-th token in all channels. Additionally, we introduce a gating mechanism and residual connections to adaptively control the refinement strength and preserve the original information to generate the discriminative frozen feature $\widetilde{\mathcal{F}}_1^{\ast}$, which can be defined as:  
\begin{equation}
\begin{split}
        &\widetilde{\mathcal{F}}_1^{\ast}=\delta \cdot \widetilde{\mathbf{T}}_{1}^{\ast} + \hat{\mathcal{F}}_1^{\ast}, \ \delta=\mathbf{Sig}(w_g\bar{{\mathcal{F}}}_1^{\ast}+b_g), \\
\end{split}
\end{equation}
where $\delta$ denotes a gating coefficient, $\bar{{\mathcal{F}}}_1^{\ast}$ represents a global mean of $\hat{\mathcal{F}}_1^{\ast}$. Subsequently, the optimized feature $\widetilde{\mathcal{F}}_1^{\ast}$ is injected into the first frozen Transformer block to obtain the feature ${\mathcal{F}}_2^{\ast}$, which is used for fine-tuning in the next block.

\textbf{Spatial-aware Expert Enhancer (SEE).} During the interactive fine-tuning process, we introduce a spatial-aware expert enhancer (SEE) to enhance and update the information in trainable features, aiming to strengthen their perception of spatial structures. Specifically, we employ three distinct branches ($i.e.$, Directional Laplacian Filter, Adaptive Max-pooling, and Multi-scale Operation) to capture spatial structural cues from the trainable features. These cues are then dynamically weighted through learnable parameters, allowing the model to selectively acquire spatial information as needed. Technically, the input feature $\tilde{{\mathcal{F}}_1^{e}}$ is restored to three spatial scales and reshaped as feature maps $\{\mathcal{F}_2^{\diamond},\mathcal{F}_3^{\diamond}, \mathcal{F}_4^{\diamond}\}$. 
In the first branch, we adopt a directional Laplacian filter to capture structural information by enhancing second-order variations along a specific spatial axis, which emphasizes high-frequency components, such as boundaries and textures. For each scale $i\in \{2,3,4\}$, the DLF-enhanced feature ${\mathcal{F}}_i^{d}$ can be defined as:
\begin{equation}
\begin{split}
        &{\mathcal{F}}_i^{d} = \sum_{p=-1}^{1} \sum_{q=-1}^{1} \mathbf{K}[p, q] \cdot \breve{\mathcal{F}}_i^{\diamond}[c, h + p, w + q], \\
\end{split}
\end{equation}
where $\mathbf{K}$$\in$ $\mathbb{R}^{3 \times 3}$ is a fixed-directional Laplacian kernel that approximates second-order derivatives. $\breve{\mathcal{F}}_i^{\diamond}$ is the input feature after reflection padding, $p, q \in \{-1, 0, 1\}$ are the relative spatial offsets. In the second branch, we apply adaptive max-pooling to extract global structural features, which often highlight salient spatial patterns and regions, $i.e.$, ${\mathcal{F}}_i^{a}$ = $\max_{(h, w) \in \Omega_i} \mathcal{F}_i^{\diamond}[c, h, w]$, where $\max_{(h, w) \in \Omega_i}$ represents adaptive max-pooling over the full spatial domain. In the third branch, we use a multi-scale operation ($\mathcal{MSO}(\cdot)$), which consists of depthwise convolutions with different kernels ($i.e.$, 3, 5, and 7), to extract local context at different scales, enhancing both fine and coarse structural patterns, $i.e.$, ${\mathcal{F}}_i^{m} = \sum_{n=1}^{3} \mathcal{MSO}_{2n+1}(\mathcal{F}_i^{\diamond})$, where $2n+1$ represents the kernel size of depthwise convolutions within the multi-scale operation. Considering the differences in structural information captured by each branch, we employ dynamic weights to control the outputs of the three branches to produce the powerful trainable feature $\tilde{{\mathcal{F}}_2^{e}}$ for subsequent fine-tuning. The progress is formulated as follows: 
\begin{equation}
\begin{split}
        &\tilde{\mathcal{F}}_2^{e}=\mathsf{Cat}[\mathsf{RS}(\tilde{\mathcal{F}}_2^{\diamond}),\mathsf{RS}(\tilde{\mathcal{F}}_3^{\diamond}),\mathsf{RS}(\tilde{\mathcal{F}}_4^{\diamond})]+\tilde{\mathcal{F}}_1^{e}, \\
        &\tilde{\mathcal{F}}_i^{\diamond}=\mathcal{F}_i^{\diamond}+{\textstyle\sum_{z\in \{d, a, m\}}^{}}w_z \cdot \mathcal{F}_i^{z}, \: \: i=2,3,4,\\
\end{split}
\end{equation}
where $w_d$, $w_a$, and $w_m$ are different learnable weights used to flexibly output the structural information extracted by different branches, and their sum is equal to 1. Similarly, in the second stage of fine-tuning, the generated trainable and frozen features $\tilde{\mathcal{F}}_2^{e}$ and $\mathcal{F}_2^{\ast}$ are used as inputs to our EC adapter, where the same operations are applied to optimize the generation of features $\tilde{\mathcal{F}}_3^{e}$ and $\mathcal{F}_3^{\ast}$. As depicted in Fig. \ref{liuchengtu}, the entire fine-tuning process lasted for four stages.

\subsection{Loss Function}
After the completion of the four fine-tuning stages, the optimized features are passed into a lightweight Transformer-based mask decoder \cite{mask2}, comprising 3.19 million trainable parameters. The WEFT method is fine-tuned under the supervision of a composite loss function that integrates binary cross-entropy ($\mathcal{L}_{bce}$) and Dice coefficient ($\mathcal{L}_{dice}$) losses, which can be formulated as follows:
\begin{equation}
\begin{split}
&\mathcal{L}_{all}=\beta\mathcal{L}_{bce}+\gamma\mathcal{L}_{dice}, \\
\end{split}
\end{equation}
where $\beta$ and $\gamma$ denote the hyperparameters set to 5 and 2.

\section{Experiments}
\subsection{Experimental Settings}
\noindent \textbf{Datasets.} We perform experiments separately on three ORSIs datasets: ORSSD \cite{ORSSD}, comprising 600 training images and 200 testing images; EORSSD \cite{EORSSD}, consisting of 1,400 training images and 600 testing images; and ORSIs-4199 \cite{ORSIs-4199}, which includes 2,000 training images and 2,199 testing images.

\noindent \textbf{Evaluation Metrics.} We employ five evaluation metrics to assess the performance of our WEFT method, including mean Intersection over Union (\textbf{mIoU}), average F-measure (\textbf{AF$_{m}$}), mean Dice coefficient (\textbf{mDice}), structural similarity measure (\textbf{S$_m$}), and mean absolute error (\textbf{MAE}).

\noindent \textbf{Implementation details.} We implement our WEFT model based on the PyTorch framework and train it using four NVIDIA RTX 4090 GPUs with 24GB of memory. During training, input images are resized to 512 $\times$ 512, with a batch size of 6 and an initial learning rate of 5e-5. The model is optimized using the AdamW optimizer over 80K iterations.

\begin{table*}[t]
\centering
\setlength{\tabcolsep}{3pt}
\renewcommand{\arraystretch}{0.9}
\resizebox*{0.97\textwidth}{65mm}{
\begin{tabular}{c|c|ccccc|ccccc|ccccc}
\hline \hline 
\multirow{2}{*}{\textbf{Methods}} & \multirow{2}{*}{\begin{tabular}[c]{@{}c@{}}\textbf{Trainable}\\ \textbf{Params (M)}\end{tabular}} & \multicolumn{5}{c|}{\textbf{ORSSD (200 images)}}                 & \multicolumn{5}{c|}{\textbf{EORSSD (600 images)}}                & \multicolumn{5}{c}{\textbf{ORSIs-4199 (2199 images)}}              \\
                         &                           & \cellcolor{magenta!12} \textbf{mIoU} $\uparrow$   & \cellcolor{magenta!12}\textbf{AF$_m$} $\uparrow$    & \cellcolor{magenta!12}\textbf{mDice} $\uparrow$ & \cellcolor{magenta!12}\textbf{S$_m$} $\uparrow$    & \cellcolor{magenta!12}\textbf{MAE} $\downarrow$   & \cellcolor{magenta!12}\textbf{mIoU} $\uparrow$   & \cellcolor{magenta!12}\textbf{AF$_m$} $\uparrow$    & \cellcolor{magenta!12}\textbf{mDice} $\uparrow$ & \cellcolor{magenta!12}\textbf{S$_m$} $\uparrow$    & \cellcolor{magenta!12}\textbf{MAE} $\downarrow$    & \cellcolor{magenta!12}\textbf{mIoU} $\uparrow$   & \cellcolor{magenta!12}\textbf{AF$_m$} $\uparrow$    & \cellcolor{magenta!12}\textbf{mDice} $\uparrow$ & \cellcolor{magenta!12}\textbf{S$_m$} $\uparrow$    & \cellcolor{magenta!12}\textbf{MAE} $\downarrow$    \\ \hline \hline 
LVNet$_{19}$                    & -                         & 0.7276 & 0.7506 & 0.7897 & 0.8730 & 0.0207 & 0.6734 & 0.6306 & 0.7167 & 0.8355 & 0.0145 & -      & -      & -      & -      & -      \\ 
VST$_{21}$                      & 44.48                     & 0.8531 & 0.8262 & 0.8895 & 0.9267 & 0.0094 & 0.8125 & 0.7089 & 0.8418 & 0.8829 & 0.0067 & 0.7809 & 0.7947 & 0.8444 & 0.8733 & 0.0281 \\  
DAFNet$_{21}$                   & 29.35                     & 0.8234 & 0.7876 & 0.8739 & 0.9119 & 0.0113 & 0.8002 & 0.6423 & 0.8297 & 0.8830 & 0.0060 & -      & -      & -      & -      & -      \\
PA-KRN$_{21}$                   & 141.06                    & 0.8382 & 0.8548 & 0.8955 & 0.9145 & 0.0139 & 0.8047 & 0.7993 & 0.8655 & 0.8802 & 0.0104 & 0.7371 & 0.8200 & 0.8167 & 0.8427 & 0.0382 \\ 
EMFINet$_{22}$                  & 95.09                     & 0.8350 & 0.8617 & 0.8864 & 0.9267 & 0.0109 & 0.8004 & 0.7984 & 0.8507 & 0.8891 & 0.0084 & 0.7598 & 0.8186 & 0.8335 & 0.8612 & 0.0330 \\ 
MCCNet$_{22}$                   & 67.65                     & 0.8554 & 0.8957 & 0.9036 & 0.9334 & 0.0087 & 0.8169 & 0.8137 & 0.8724 & 0.8954 & 0.0066 & 0.7768 & 0.8592 & 0.8475 & 0.8682 & 0.0316 \\ 
MJRBM$_{22}$                    & 43.54                     & 0.8169 & 0.8022 & 0.8529 & 0.9097 & 0.0163 & 0.7931 & 0.7066 & 0.8217 & 0.8786 & 0.0099 & 0.7466 & 0.7995 & 0.8128 & 0.8530 & 0.0374 \\ 
ERPNet$_{22}$                   & 56.48                     & 0.8224 & 0.8356 & 0.8687 & 0.9153 & 0.0135 & 0.7887 & 0.7554 & 0.8273 & 0.8812 & 0.0089 & 0.7533 & 0.8024 & 0.8224 & 0.8606 & 0.0357 \\ 
ESGNet$_{23}$        & -                         & 0.8380 & 0.8860 & 0.8929 & 0.9243 & 0.0098 & 0.8023 & {0.8557} & 0.8624 & 0.8880 & 0.0070 & 0.7818 & \textbf{\color{myblue}0.8740} & 0.8580 & 0.8730 & 0.0289 \\ 
GeleNet$_{23}$        & 25.50                     & 0.8575 & \textbf{\color{myblue}0.9038} & 0.9070 & 0.9326 & 0.0080 & 0.8017 & 0.8528 & 0.8669 & 0.8894 & \textbf{\color{myblue}0.0055} & 0.7885 & 0.8681 & 0.8614 & \textbf{\color{myblue}0.8766} & \textbf{\color{myblue}0.0264} \\ 
ICON-P$_{23}$                     & 65.68                     & 0.8403 & 0.8444 & 0.8964 & 0.9162 & 0.0116 & 0.8122 & 0.8065 & 0.8792 & 0.8821 & 0.0073 & 0.7788 & 0.8531 & 0.8523 & 0.8692 & 0.0282 \\ 
ACCoNet$_{23}$                  & 102.55                    & 0.8539 & 0.8806 & 0.9026 & 0.9335 & 0.0088 & 0.8118 & 0.7969 & 0.8651 & 0.8898 & 0.0074 & 0.7778 & 0.8581 & 0.8484 & 0.8711 & 0.0314 \\ 
SRAL$_{23}$                     & 31.90                     & 0.8467 & 0.8514 & 0.9020 & 0.9225 & 0.0107 & 0.8211 & 0.8146 & \textbf{\color{myblue}0.8824} & 0.8869 & 0.0069 & 0.7705 & 0.8230 & 0.8451 & 0.8678 & 0.0307 \\ 
TLCKDNet$_{24}$                 & 52.09                     & \textbf{\color{myblue}0.8689} & 0.8719 & \textbf{\color{myblue}0.9142} & 0.9316 & 0.0082 & \textbf{\color{myTeal}0.8380} & 0.7969 & \textbf{\color{myTeal}0.8895} & 0.8955 & 0.0056 & -      & -      & -      & -      & -      \\ 
SOLNet$_{24}$    & \textbf{\color{reda}6.50}                      & 0.8302 & 0.8925 & 0.8885 & 0.9195 & 0.0111 & 0.7792 & 0.8392 & 0.8467 & 0.8814 & 0.0078 & -      & -      & -      & -      & -      \\
UDCNet$_{24}$         & 72.20                     & 0.8680 & 0.8932 & 0.9131 & \textbf{\color{myTeal}0.9381} & \textbf{\color{myblue}0.0068} & 0.8073 & 0.8211 & 0.8697 & {0.8957} & 0.0056 & \textbf{\color{myblue}0.7913} & {0.8648} & \textbf{\color{myblue}0.8640} & 0.8744 & 0.0266 \\ 
SFANet$_{24}$         & 25.10                     & 0.8593 & 0.8984 & 0.9077 & 0.9345 & 0.0077 & 0.8136 & 0.8492 & 0.8720 & 0.8955 & 0.0058 & 0.7774 & 0.8647 & 0.8509 & 0.8703 & 0.0292 \\ 
ADSTNet$_{24}$        & 62.09                     & 0.8459 & 0.8979 & 0.8980 & 0.9296 & 0.0086 & 0.8046 & 0.8532 & 0.8651 & 0.8912 & 0.0065 & 0.7693 & 0.8655 & 0.8460 & 0.8647 & 0.0318 \\ 
BCARNet$_{25}$          & \textbf{\color{myblue}24.00}                     & 0.8600 & \textbf{\color{myTeal}0.9073} & 0.9088 & \textbf{\color{myblue}0.9361} & 0.0071 & 0.8248 & \textbf{\color{myTeal}0.8740} & 0.8821 & \textbf{\color{myblue}0.8969} & \textbf{\color{myTeal}0.0054} & 0.7795 & {0.8666} & 0.8502 & {0.8694} & 0.0306 \\ 
LGIPNet$_{25}$               & 65.80                     & 0.8572 & 0.8924 & 0.9051 & 0.9348 & 0.0082 & 0.8154 & \textbf{\color{myblue}0.8560} & 0.8721 & 0.8938 & 0.0065 & 0.7899 & 0.8280 & 0.8583 & 0.8754 & 0.0288 \\ 
DPU-Former$_{25}$         & 44.20                     & \textbf{\color{myTeal}0.8728} & 0.9024 & \textbf{\color{myTeal}0.9163} & 0.9312 & \textbf{\color{myTeal}0.0062} & \textbf{\color{myblue}0.8268} & 0.8461 & 0.8811 & \textbf{\color{reda}0.9011} & 0.0056 & \textbf{\color{myTeal}0.7961} & \textbf{\color{myTeal}0.8816} & \textbf{\color{myTeal}0.8677} & \textbf{\color{myTeal}0.8769} & \textbf{\color{myTeal}0.0263} \\ 
\cellcolor{myOrange!12} \textbf{Ours}                  & \cellcolor{myOrange!12}\textbf{\color{myTeal}14.37}                     & \cellcolor{myOrange!12}\textbf{\color{reda}0.8964} & \cellcolor{myOrange!12}\textbf{\color{reda}0.9213} & \cellcolor{myOrange!12}\textbf{\color{reda}0.9394} & \cellcolor{myOrange!12}\textbf{\color{reda}0.9383} & \cellcolor{myOrange!12}\textbf{\color{reda}0.0056} & \cellcolor{myOrange!12}\textbf{\color{reda}0.8621} & \cellcolor{myOrange!12}\textbf{\color{reda}0.8810} & \cellcolor{myOrange!12}\textbf{\color{reda}0.9188} & \cellcolor{myOrange!12}\textbf{\color{myTeal}0.9006} & \cellcolor{myOrange!12}\textbf{\color{reda}0.0048} & \cellcolor{myOrange!12}\textbf{\color{reda}0.7999} & \cellcolor{myOrange!12}\textbf{\color{reda}0.8826} & \cellcolor{myOrange!12}\textbf{\color{reda}0.8696} & \cellcolor{myOrange!12}\textbf{\color{reda}0.8772} & \cellcolor{myOrange!12}\textbf{\color{reda}0.0238} \\ \hline  \hline 
\end{tabular}}
\caption{Comparison with 21 state-of-the-arts (SOTA) methods on three widely-utilized ORSIs datasets. The top three results are highlighted in \textbf{\color{reda} red}, \textbf{\color{myTeal} green}, and \textbf{\color{myblue} blue}. ``$\uparrow$'' and ``$\downarrow$'' respectively indicate higher-is-better and lower-is-better performance.}
\label{Table-rssod}
\end{table*}

\begin{table*}[t]
\centering
\setlength{\tabcolsep}{1pt}
\begin{minipage}{0.39\textwidth}
\centering
\resizebox{\textwidth}{11mm}{
\begin{tabular}{c|c|ccc|ccc|ccc}
\hline \hline
\multirow{2}{*}{\textbf{Methods}} & \multirow{2}{*}{\begin{tabular}[c]{@{}c@{}}\textbf{Trainable}\\ \textbf{Params (M)}\end{tabular}} & \multicolumn{3}{c|}{\textbf{CAMO (250 images)}} & \multicolumn{3}{c|}{\textbf{COD10K (2026 images)}} & \multicolumn{3}{c}{\textbf{NC4K (4121 images)}} \\
                         &                           & \cellcolor{magenta!12} \textbf{mIoU} $\uparrow$   & \cellcolor{magenta!12}\textbf{AF$_m$} $\uparrow$    & \cellcolor{magenta!12}\textbf{mDice} $\uparrow$  & \cellcolor{magenta!12}\textbf{mIoU} $\uparrow$   & \cellcolor{magenta!12}\textbf{AF$_m$} $\uparrow$    & \cellcolor{magenta!12}\textbf{mDice} $\uparrow$   & \cellcolor{magenta!12}\textbf{mIoU} $\uparrow$   & \cellcolor{magenta!12}\textbf{AF$_m$} $\uparrow$    & \cellcolor{magenta!12}\textbf{mDice} $\uparrow$  \\ \hline \hline
FSPNet$_{23}$                   & 273.79                    & 0.7367  & 0.8293 & 0.8186 & 0.6771  & 0.7364  & 0.7579  & 0.7612 & 0.8258 & 0.8308 \\ 
VSCode$_{24}$                   & 117.41                    & 0.7730  & 0.8431 & 0.8463 & 0.7264  & 0.7836  & 0.8046  & 0.7927 & 0.8576 & 0.8573 \\ 
FSEL$_{24}$                     & 67.13                     & \textbf{\color{myblue}0.7996}  & \textbf{\color{myblue}0.8641} & \textbf{\color{myTeal}0.8742} & \textbf{\color{myblue}0.7440}  & 0.7962  & \textbf{\color{myblue}0.8243}  & \textbf{\color{myblue}0.7997} & \textbf{\color{myblue}0.8636} & \textbf{\color{myblue}0.8683} \\ 
ZoomXNet$_{24}$                 & 84.78                     & \textbf{\color{myTeal}0.8090}  & \textbf{\color{myTeal}0.8694} & \textbf{\color{myblue}0.8727} & \textbf{\color{myTeal}0.7795}  & \textbf{\color{myTeal}0.8131}  & \textbf{\color{myTeal}0.8429}  & \textbf{\color{myTeal}0.8141} & \textbf{\color{myTeal}0.8715} & \textbf{\color{myTeal}0.8741} \\ 
RUN$_{25}$                      & -                         & 0.6712  & 0.7837 & 0.7661 & 0.6500  & 0.7383  & 0.7451  & 0.7249 & 0.8208 & 0.8095 \\ 
DPU-Former$_{25}$               & \textbf{\color{myblue}44.20}                     & 0.7814  & 0.8578 & 0.8582 & 0.7394  & \textbf{\color{myblue}0.8075}  & 0.8175  & 0.7904 & 0.8629 & 0.8582 \\ 
\cellcolor{myOrange!10} \textbf{Ours}                  & \cellcolor{myOrange!10}\textbf{\color{reda}14.37}                     & \cellcolor{myOrange!10}\textbf{\color{reda}0.8308}  & \cellcolor{myOrange!10}\textbf{\color{reda}0.8896} & \cellcolor{myOrange!10}\textbf{\color{reda}0.8972} & \cellcolor{myOrange!10}\textbf{\color{reda}0.7984}  & \cellcolor{myOrange!10}\textbf{\color{reda}0.8726}  & \cellcolor{myOrange!10}\textbf{\color{reda}0.8707}  & \cellcolor{myOrange!10}\textbf{\color{reda}0.8362} & \cellcolor{myOrange!10}\textbf{\color{reda}0.9006} & \cellcolor{myOrange!10}\textbf{\color{reda}0.8964} \\ \hline \hline
\end{tabular}}
\end{minipage}
\hfill
\begin{minipage}{0.30\textwidth}
\centering
\resizebox{\textwidth}{11mm}{
\begin{tabular}{c|c|ccc|ccc}
\hline \hline
\multirow{2}{*}{\textbf{Methods}} & \multirow{2}{*}{\begin{tabular}[c]{@{}c@{}}\textbf{Trainable}\\ \textbf{Params (M)}\end{tabular}} & \multicolumn{3}{c|}{\textbf{PASCAL-S (850 images)}} & \multicolumn{3}{c}{\textbf{HKU-IS (4447 images)}}\\
                         &                             & \cellcolor{magenta!12} \textbf{mIoU} $\uparrow$   & \cellcolor{magenta!12}\textbf{AF$_m$} $\uparrow$    & \cellcolor{magenta!12}\textbf{mDice} $\uparrow$  & \cellcolor{magenta!12}\textbf{mIoU} $\uparrow$   & \cellcolor{magenta!12}\textbf{AF$_m$} $\uparrow$    & \cellcolor{magenta!12}\textbf{mDice} $\uparrow$     \\ \hline \hline
ICON-P$_{23}$                   & 65.68                      & 0.8112   & 0.8542   & 0.8730  & 0.8924  & 0.9255  & 0.9332  \\ 
VSCode$_{24}$                   & 117.41                     & 0.8177   & 0.8644   & 0.8744  & \textbf{\color{myblue}0.9004}  & 0.9326  & 0.9359 \\ 
MDSAM$_{24}$                    & \textbf{\color{myTeal}16.78}                          & 0.8138   & 0.8609   & 0.8721  & \textbf{\color{myTeal}0.9033}  & \textbf{\color{myTeal}0.9399}  & \textbf{\color{myTeal}0.9400}  \\ 
VST++$_{24}$                    & 112.23                     & \textbf{\color{myTeal}0.8232}   & \textbf{\color{myblue}0.8648}   & \textbf{\color{myblue}0.8782}  & -       & -       & -      \\ 
FSEL$_{24}$                     & 67.13                      & \textbf{\color{myblue}0.8227}   & \textbf{\color{myTeal}0.8690}   & \textbf{\color{myTeal}0.8812}  & 0.9000  & \textbf{\color{myblue}0.9362}  & \textbf{\color{myblue}0.9399} \\ 
DPU-Former$_{25}$               & \textbf{\color{myblue}44.20}                      & 0.8093   & 0.8621   & 0.8709  & 0.8921  & 0.9323  & 0.9336  \\ 
\cellcolor{myOrange!10} \textbf{Ours}                  & \cellcolor{myOrange!10}\textbf{\color{reda}14.37}                      & \cellcolor{myOrange!10}\textbf{\color{reda}0.8359}   & \cellcolor{myOrange!10}\textbf{\color{reda}0.8774}   & \cellcolor{myOrange!10}\textbf{\color{reda}0.8923}  & \cellcolor{myOrange!10}\textbf{\color{reda}0.9140}  & \cellcolor{myOrange!10}\textbf{\color{reda}0.9483}  & \cellcolor{myOrange!10}\textbf{\color{reda}0.9510} \\ \hline \hline
\end{tabular}}
\end{minipage}
\hfill
\begin{minipage}{0.30\textwidth}
\centering
\resizebox{\textwidth}{11mm}{
\begin{tabular}{c|c|ccc|ccc}
\hline \hline
\multirow{2}{*}{\textbf{Methods}} & \multirow{2}{*}{\begin{tabular}[c]{@{}c@{}}\textbf{Trainable}\\ \textbf{Params (M)}\end{tabular}} & \multicolumn{3}{c|}{\textbf{CVC-300 (62 images)}} & \multicolumn{3}{c}{\textbf{Kvasir (100 images)}}\\
                        &                            & \cellcolor{magenta!12} \textbf{mIoU} $\uparrow$   & \cellcolor{magenta!12}\textbf{AF$_m$} $\uparrow$    & \cellcolor{magenta!12}\textbf{mDice} $\uparrow$  & \cellcolor{magenta!12}\textbf{mIoU} $\uparrow$   & \cellcolor{magenta!12}\textbf{AF$_m$} $\uparrow$    & \cellcolor{magenta!12}\textbf{mDice} $\uparrow$  \\ \hline \hline
DCRNet$_{22}$                  & \textbf{\color{myblue}28.73}                      & 0.7974   & 0.8029  & 0.8580  & 0.8384  & 0.8814  & 0.8881  \\ 
CFANet$_{23}$                  & \textbf{\color{myTeal}25.24}                      & \textbf{\color{myblue}0.8320}   & 0.8599  & \textbf{\color{myblue}0.8958}  & \textbf{\color{myTeal}0.8698}  & 0.9053  & \textbf{\color{myTeal}0.9170} \\ 
FSEL$_{24}$                    & 67.13                      & 0.8176   & 0.8432  & 0.8828  & 0.8558  & 0.9081   & {0.8054}     \\ 
LSSNet$_{24}$                  & 35.94                      & 0.8154   & 0.8256  & 0.8835  & \textbf{\color{myblue}0.8660}  & 0.9079  & 0.9111   \\ 
MEGANet$_{24}$                 & 44.19                      & 0.8214   & \textbf{\color{myblue}0.8671}  & 0.8899  & 0.8619  & \textbf{\color{myblue}0.9213}  & \textbf{\color{myblue}0.9135}  \\ 
DPU-Former$_{25}$              & 44.20                      & \textbf{\color{myTeal}0.8414}   &\textbf{\color{myTeal}0.8764}  & \textbf{\color{myTeal}0.9024}  & 0.8609  & \textbf{\color{myTeal}0.9270}  & 0.9123 \\ 
\cellcolor{myOrange!10} \textbf{Ours}                & \cellcolor{myOrange!10}\textbf{\color{reda}14.37}                      & \cellcolor{myOrange!10}\textbf{\color{reda}0.8502}   &\cellcolor{myOrange!10} \textbf{\color{reda}0.8962}  & \cellcolor{myOrange!10}\textbf{\color{reda}0.9121}  & \cellcolor{myOrange!10}\textbf{\color{reda}0.8875}  & \cellcolor{myOrange!10}\textbf{\color{reda}0.9419}  & \cellcolor{myOrange!10}\textbf{\color{reda}0.9329}   \\ \hline \hline
\end{tabular}}
\end{minipage}
\hfill
\caption{Extended applications in \textbf{camouflage}, \textbf{natural}, and \textbf{medical} scenarios. Quantitative results with 13 state-of-the-arts segmentation models on three camouflaged object detection, two salient object detection, and two polyp segmentation datasets.}
\label{Table-extend-application}
\end{table*}

\subsection{Comparison with State-of-the-Arts}
We conduct comprehensive comparisons between our WEFT and 21 ORSIs object segmentation approaches, including LVNet\nocite{ORSSD}, VST\nocite{VST}, DAFNet\nocite{EORSSD}, PA-KRN\nocite{PAKRN}, EMFINet\nocite{EMFINet}, MCCNet\nocite{MCCNet}, MJRBM\nocite{ORSIs-4199}, ERPNet\nocite{ERPNet}, ESGNet\nocite{ESGNet}, GeleNet\nocite{GeleNet}, ICON\nocite{ICON-p}, ACCoNet\nocite{ACCorNet}, SRAL\nocite{SRAL}, TLCKDNet\nocite{TLKCDNet}, SOLNet\nocite{SOLNet}, UDCNet\nocite{UDCNet}, SFANet\nocite{SFANet}, ADSTNet\nocite{ADSTNet}, BCARNet\nocite{BCARNet}, LGIPNet\nocite{LGIPNet}, and DPU-Former\nocite{DPU-Former}. To ensure fairness, all predicted results are obtained directly from the authors or reproduced using available open-source code.

\textbf{Quantitative results.} Table \ref{Table-rssod} summarizes the performance of our WEFT model compared to 21 SOTA methods. Specifically, our ``\textbf{mIoU}'' and ``\textbf{mDice}'' metrics outperform the second-best method by substantial margins 2.70\% and 2.52\% on the ORSSD dataset and 2.88\% and 3.29\% on the EORSSD dataset. Moreover, under ``\textbf{MAE}'' metric, our method surpasses the second-best approach by 10.71\%, 12.50\%, and 10.50\% on three datasets, respectively. Likewise, other indicators exhibit considerable competitive advantages. In addition, the proposed WEFT model shows a clear advantage in terms of trainable parameters. The entire architecture contains only \textbf{14.37 M trainable parameters}, which facilitates the training and deployment of large-scale models in ORSIs tasks. Overall, the quantitative results confirm the effectiveness of our model in achieving high accuracy while maintaining computational efficiency.

\begin{figure}[t]
	\centering\includegraphics[width=0.47\textwidth,height=4.5cm]{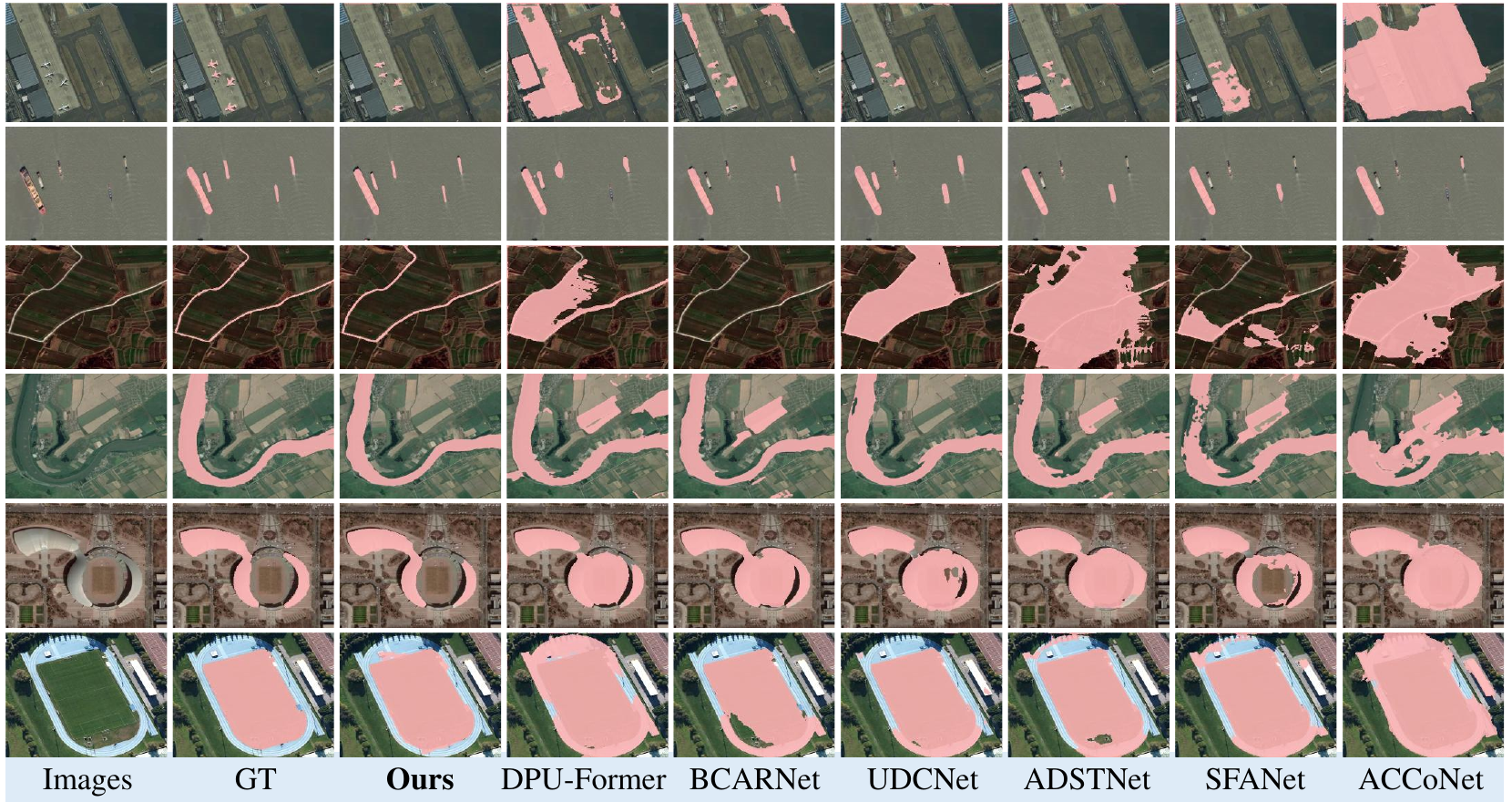}
	\captionsetup{font={small}}
	\caption{Visual results of our WEFT and existing approaches.}
	\label{Fig-visual_results}
\end{figure}

\textbf{Qualitative results.} Fig. \ref{Fig-visual_results} illustrates the visual segmentation results in various remote sensing scenarios, including aircraft ($1^{st}$ rows), ships ($2^{nd}$ row), highway ($3^{rd}$ row), river ($4^{th}$ row), building ($5^{th}$ row) and court ($6^{th}$ row). As shown in Fig. \ref{Fig-visual_results}, thanks to the efficient fine-tuning of large-scale models, our method achieves more accurate and complete segmentation of remote sensing targets compared to recent methods ($e.g.$, DPU-Former\cite{DPU-Former}, BCARNet \cite{BCARNet} and SFANet \cite{SFANet}).

\subsection{Extended Applications}
To further validate the generalizability of our WEFT framework, we follow the same settings to conduct additional experiments on camouflaged object detection (COD), salient object detection (SOD), and polyp segmentation (PS) tasks. Table \ref{Table-extend-application} compares our method with multiple existing methods \nocite{FSPNet,ICON-p, DCRNet, CFANet, LSSNet, MEGANet,VSCode,FSEL,ZoomXNet, MDSAM, VST++,DPU-Former,RUN} on 7 segmentation datasets \nocite{CAMO,COD10K,NC4K,PASCAL-S,HKU,CVC300,kvasir}. In Table \ref{Table-extend-application}, it is evident that our WEFT achieves outstanding performance across various image types, which can be attributed to the strong modeling capacity of large-scale foundation models.

\subsection{Ablation Study}

\begin{figure}[t]
	\centering\includegraphics[width=0.47\textwidth,height=2.7cm]{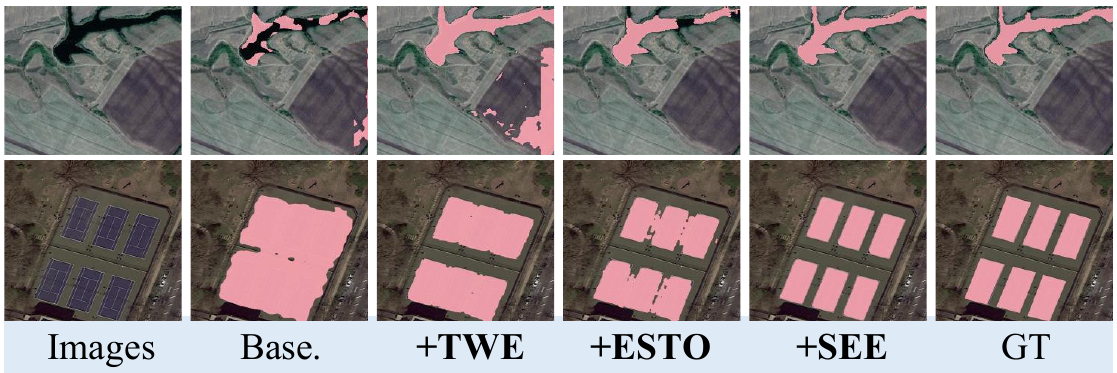}
	\captionsetup{font={small}, justification=raggedright}
	\caption{Visual results of the effectiveness of each component.}
	\label{AS-visual_results}
\end{figure}

\textbf{Effect of each component.} Table \ref{component_AS} presents the quantitative results for each individual component of our WEFT framework. To be specific, ``\textbf{Base.}'' (Table \ref{component_AS}(a)) consists of a frozen UniPerceiver-L network, combined with a trainable mask decoder. Table \ref{component_AS}(b) shows the effectiveness of our ``\textbf{TWE}'' extractor, demonstrating that integrating dynamic wavelet experts with knowledge guidance improves the adaptability of frozen frameworks to ORSIs tasks. Moreover, as illustrated in Table \ref{component_AS} (c) and (e), ``\textbf{ESTO}'' yields notable improvements in segmentation precision by enabling subspace token refinement and boundary-aware enhancement. Furthermore, we integrate the designed ``\textbf{SEE}'' into the WEFT (as shown in Table \ref{component_AS} (d), (f), and (g)), which further strengthens the spatial perception of experts during the fine-tuning process, thereby improving the performance. Additionally, in Fig. \ref{AS-visual_results}, we provide visualized results obtained by progressively incorporating each component ($i.e.$, TWE, ESTO, and SEE), illustrating that the predicted results increasingly resemble the ground truth. In conclusion, each component is essential and collectively contributes to improvements over the baseline by 12.09\% and 11.27\% in terms of the ``\textbf{mIoU}'' metric on two ORSIs datasets.

\begin{table}[t]
\centering
\setlength{\tabcolsep}{1pt}
\renewcommand{\arraystretch}{0.9}
\resizebox*{0.48\textwidth}{28mm}{
\begin{tabular}{c|cccc|ccc|ccc}
\toprule[1pt]
\multirow{2}{*}{\textbf{Num.}} & \multicolumn{4}{c|}{\textbf{Component Setting}} & \multicolumn{3}{c|}{\textbf{EORSSD (600 images)}} & \multicolumn{3}{c}{\textbf{ORSIs-4199 (2199 images)}} \\
                      & \cellcolor{myblue!12} \textbf{Base.}  & \cellcolor{myblue!12}\textbf{TWE}  & \cellcolor{myblue!12}\textbf{ESTO}  & \cellcolor{myblue!12}\textbf{SEE}  & \cellcolor{magenta!12} \textbf{mIoU} $\uparrow$   & \cellcolor{magenta!12}\textbf{AF$_m$} $\uparrow$    & \cellcolor{magenta!12}\textbf{mDice} $\uparrow$  & \cellcolor{magenta!12}\textbf{mIoU} $\uparrow$   & \cellcolor{magenta!12}\textbf{AF$_m$} $\uparrow$    & \cellcolor{magenta!12}\textbf{mDice} $\uparrow$   \\ \midrule[1pt]
(a)                     & $\checkmark$      &          &          &          & 0.7691  & 0.8193  & 0.8521  & 0.7189   & 0.8153   & 0.8096  \\ 
(b)                     & $\checkmark$      & $\checkmark$        &          &          & 0.8253  & 0.8604  & 0.8934  & 0.7528   & 0.8441   & 0.8373  \\ 
(c)                     & $\checkmark$      &          & $\checkmark$        &          & 0.8196  & 0.8637  & 0.8864  & 0.7715   & 0.8596   & 0.8481  \\ 
(d)                     & $\checkmark$      &          &          & $\checkmark$        & 0.8048  & 0.8411  & 0.8785  & 0.7451   & 0.8403   & 0.8310  \\ 
(e)                     & $\checkmark$      & $\checkmark$        & $\checkmark$        &          & 0.8323  & 0.8668  & 0.8949  & 0.7864   & 0.8764   & 0.8608  \\ 
(f)                     &  $\checkmark$     &          & $\checkmark$        & $\checkmark$        & 0.8312  & 0.8579  & 0.8964  & 0.7902   & 0.8705   & 0.8635  \\
(g)                     & $\checkmark$      & $\checkmark$        &          & $\checkmark$        & 0.8275  & 0.8597  & 0.8937  & 0.7939   & 0.8756   & 0.8649  \\ 
\cellcolor{myOrange!10} (h)                     & \cellcolor{myOrange!10} $\checkmark$      & \cellcolor{myOrange!10} $\checkmark$        & \cellcolor{myOrange!10} $\checkmark$        & \cellcolor{myOrange!10} $\checkmark$        & \cellcolor{myOrange!10} \textbf{\color{reda}0.8621}  & \cellcolor{myOrange!10} \textbf{\color{reda}0.8810}  & \cellcolor{myOrange!10} \textbf{\color{reda}0.9188}  & \cellcolor{myOrange!10} \textbf{\color{reda}0.7999}   & \cellcolor{myOrange!10} \textbf{\color{reda}0.8826}   & \cellcolor{myOrange!10} \textbf{\color{reda}0.8696}  \\ \bottomrule[1pt]
\end{tabular}}
\caption{Ablation analysis of individual components in our WEFT framework on EORSSD and ORSIs-4199 datasets.}
\label{component_AS}
\end{table}

\begin{table}[t]
\centering

\begin{minipage}{\linewidth}
\centering
\setlength{\tabcolsep}{4pt}
\renewcommand{\arraystretch}{1}
\resizebox*{\linewidth}{22mm}{
\begin{tabular}{c|c|ccc|ccc}
\toprule[1pt]
\multirow{2}{*}{\textbf{Num.}} & \multirow{2}{*}{\begin{tabular}[c]{@{}c@{}}\textbf{Expert}\\ \textbf{Allocation}\end{tabular}} & \multicolumn{3}{c|}{\textbf{EORSSD (600 images)}} & \multicolumn{3}{c}{\textbf{ORSIs-4199 (2199 images)}} \\
                      &                                                                              & \cellcolor{magenta!12} \textbf{mIoU} $\uparrow$   & \cellcolor{magenta!12}\textbf{AF$_m$} $\uparrow$    & \cellcolor{magenta!12}\textbf{mDice} $\uparrow$  & \cellcolor{magenta!12}\textbf{mIoU} $\uparrow$   & \cellcolor{magenta!12}\textbf{AF$_m$} $\uparrow$    & \cellcolor{magenta!12}\textbf{mDice} $\uparrow$   \\ \midrule[1pt]
(a) & Exp.=1  & 0.7951  & 0.8464  & 0.8726  & 0.7278   & 0.8281   & 0.8156  \\ 
(b) & Exp.=2  & 0.8039  & 0.8501  & 0.8799  & 0.7499   & \textbf{\color{reda}0.8491}   & 0.8345  \\ 
(c) & Exp.=6  & 0.8138  & 0.8501  & {0.8867}  & \textbf{\color{reda}0.7544}   & 0.8437   & 0.8371  \\ 
\cellcolor{myOrange!10} (d) & \cellcolor{myOrange!10}Exp.=4  & \cellcolor{myOrange!10}\textbf{\color{reda}0.8253}  & \cellcolor{myOrange!10}\textbf{\color{reda}0.8604}  & \cellcolor{myOrange!10}\textbf{\color{reda}0.8934}  & \cellcolor{myOrange!10}{0.7528}   & \cellcolor{myOrange!10}{0.8441}   & \cellcolor{myOrange!10}\textbf{\color{reda}0.8373}  \\ 
\bottomrule[1pt]
\end{tabular}}
\captionof{table}{Ablation analysis of the number of wavelet experts.}
\label{WE_number}
\end{minipage}

\begin{minipage}{\linewidth}
\centering
\setlength{\tabcolsep}{4pt}
\renewcommand{\arraystretch}{1}
\resizebox*{\linewidth}{22mm}{
\begin{tabular}{c|c|ccc|ccc}
\toprule[1pt]
\multirow{2}{*}{\textbf{Num.}} & \multirow{2}{*}{\begin{tabular}[c]{@{}c@{}}\textbf{Subspace}\\ \textbf{Setting}\end{tabular}} & \multicolumn{3}{c|}{\textbf{EORSSD (600 images)}} & \multicolumn{3}{c}{\textbf{ORSIs-4199 (2199 images)}} \\
                       &                                                                              & \cellcolor{magenta!12} \textbf{mIoU} $\uparrow$   &\cellcolor{magenta!12} \textbf{AF$_m$} $\uparrow$    & \cellcolor{magenta!12}\textbf{mDice} $\uparrow$  & \cellcolor{magenta!12}\textbf{mIoU} $\uparrow$   & \cellcolor{magenta!12}\textbf{AF$_m$} $\uparrow$    & \cellcolor{magenta!12}\textbf{mDice} $\uparrow$   \\ \midrule[1pt]
(a) & H = 2  & 0.8129  & 0.8541  & 0.8838  & 0.7705   & 0.8567   & \textbf{\color{reda}0.8484}  \\ 
(b) & H = 8  & 0.8130  & 0.8588  & 0.8832  & 0.7628   & 0.8520   & 0.8436  \\ 
(c) & H = 16 & 0.8124  & 0.8559  & 0.8828  & 0.7634   & 0.8533   & 0.8445  \\ 
\cellcolor{myOrange!10} (d) & \cellcolor{myOrange!10} H = 4 & \cellcolor{myOrange!10} \textbf{\color{reda}0.8196}  & \cellcolor{myOrange!10} \textbf{\color{reda}0.8637}  & \cellcolor{myOrange!10} \textbf{\color{reda}0.8864}  & \cellcolor{myOrange!10} \textbf{\color{reda}0.7715}   & \cellcolor{myOrange!10} \textbf{\color{reda}0.8596}   & \cellcolor{myOrange!10} 0.8481  \\ 
\bottomrule[1pt]
\end{tabular}}

\captionof{table}{Ablation analysis of the number of subspaces.}
\label{Sub_number}
\end{minipage}
\end{table}

\begin{figure}[t]
	\centering\includegraphics[width=0.46\textwidth,height=3.3cm]{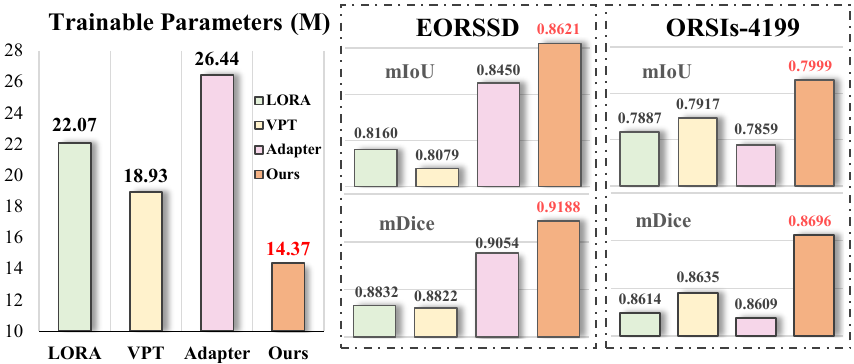}
	\captionsetup{font={small}, justification=raggedright}
	\caption{Comparison of different fine-tuning strategies.}
	\label{FT_Strategy}
\end{figure}

\textbf{Effect of expert allocation.} We perform an ablation analysis on the number of experts within our ``\textbf{TWE}'' extractor. The results from Table \ref{WE_number} (a)–(d) indicate that employing multiple wavelet experts significantly outperforms the use of a single wavelet expert. However, the lower performance with six experts compared to four suggests that not all experts contribute positively during fine-tuning, indirectly proving the value of our TER strategy. Accordingly, we select four wavelet experts for the final configuration.

\textbf{Effect of subspace setting.} In Table \ref{Sub_number}, we present the experimental results obtained using different numbers of subspaces in the proposed ``\textbf{ESTO}'' component. It can be observed that optimal performance is achieved when the number of subspaces is set to four. Based on this, we adopt four subspaces in our subsequent experimental settings.

\textbf{Effect of fine-tuning strategy.} Fig. \ref{FT_Strategy} provides a detailed comparative analysis of different fine-tuning strategies, with all experiments conducted under the same framework. These results demonstrate that, compared to the LORA \cite{LORA}, VPT \cite{VPT}, and Adapter \cite{ViT-adapter} strategies, our WEFT not only significantly reduces trainable parameters, but also achieves superior performance on ORSIs object segmentation. This advantage is attributed to the lightweight and efficient design of all its components.


\section{Conclusions}
In this paper, we propose a novel fine-tuning strategy, called WEFT (Wavelet Expert-guided Fine-Tuning), which efficiently adapts frozen large-scale models to ORSIs segmentation tasks. First, we design a TWE extractor that models wavelet experts enriched with task-specific knowledge, providing effective support for fine-tuning. Second, we construct an EC adapter that enables efficient integration between frozen and trainable features, allowing both types of information to be updated. Extensive experiments demonstrate that our WEFT method significantly outperforms 35 SOTA models across 10 object segmentation datasets.

\section*{Acknowledgments} This work was supported in part by the National Science Fund of China (No. 62276135, U24A20330 and 62361166670).


\bibliography{aaai2026}

\end{document}